\documentclass[11pt,a4paper]{article}
\usepackage[hyperref]{acl2019}
\usepackage{times}
\usepackage{latexsym}
\usepackage{amsmath}
\usepackage[acronym,shortcuts,acronymlists={hidden}]{glossaries}
\usepackage{multirow,array, graphicx}
\usepackage{enumerate}

\glsdisablehyper
\usepackage{url}

\newacronym{nmt}{NMT}{Neural Machine Translation}
\newacronym{ape}{APE}{Automatic Post Editing}
\newacronym{mt}{MT}{Machine Translation}

\aclfinalcopy % Uncomment this line for the final submission
\title{Unbabel's Submission to the WMT2019 APE Shared Task:\\BERT-based Encoder-Decoder for Automatic Post-Editing}
\author{Ant\'{o}nio V. Lopes \\
  Unbabel \\ 
  \And
  M. Amin Farajian \\
   Unbabel \\ 
  \And
  Gon\c{c}alo M. Correia \\
  Instituto de Telecomunica\c{c}\~oes \\
  \AND
  Jonay Trenous \\
   Unbabel \\ 
  \And
  Andr\'{e} F. T. Martins \\
   Unbabel \\ 
  \AND \\
  \texttt{\{antonio.lopes, amin, sony,andre.martins\}@unbabel.com}\\
  \texttt{goncalo.correia@lx.it.pt}
  }

\date{}

\begin{document}
\maketitle
\begin{abstract}
This paper describes Unbabel's submission to the WMT2019 APE Shared Task for the English-German language pair.  
Following the recent rise of large, powerful, pre-trained models, we adapt the BERT pretrained model to perform Automatic Post-Editing in an encoder-decoder framework. Analogously to dual-encoder architectures we develop a BERT-based encoder-decoder (BED) model in which a single pretrained BERT encoder receives both the source \texttt{src} and machine translation \texttt{mt} strings.
Furthermore, we explore a conservativeness factor to constrain the \acrshort{ape} system to perform fewer edits. 
As the official results show, when trained on a weighted combination of in-domain and artificial training data, our BED system with the conservativeness penalty improves significantly the translations of a strong \ac{nmt} system by $-0.78$ and $+1.23$ in terms of TER and BLEU, respectively. Finally, our submission achieves a new state-of-the-art, ex-aequo, in English-German APE of NMT.

\end{abstract}

\section{Introduction}\label{intro}

\ac{ape}  aims to improve the quality of an existing \ac{mt} system by learning from human edited samples.
It first started by the automatic article selection for English noun phrases \cite{Knight:1994:APD:2891730.2891850} and continued by correcting the errors of more complex statistical \acrshort{mt} systems \cite{wmt-findings:2015, wmt-findings:2016, wmt-findings:2017}.
In 2018, the organizers of the WMT shared task introduced, for the first time, the automatic post-editing of neural MT systems \cite{wmt-findings:2018}.

Despite its successful application to SMT systems, it has been more challenging to automatically post edit the strong~\ac{nmt} systems \cite{DBLP:conf/wmt/Junczys-Dowmunt18a}.
This mostly is due to the fact that  high quality~\acrshort{nmt} systems make fewer mistakes, limiting the improvements obtained by  state-of-the-art APE systems such as self-attentive transformer-based models  \cite{tebbifakhr-etal-2018-multi, DBLP:conf/wmt/Junczys-Dowmunt18a}.
In spite of these findings %\cite{wmt-findings:2018} 
and considering the dominance of the~\acrshort{nmt} approach in both the academic and industrial applications, the WMT shared task organizers decided to move completely to the NMT paradigm this year and ignore the SMT technology. They also provide the previous year in-domain training set (i.e. $13k$ of \texttt{<src,mt,pe>} triplets) further increasing the difficulty of the task.

Training state-of-the-art \ac{ape} systems capable of improving high quality \acrshort{nmt} outputs requires large amounts of training data, which is not always available, in particular for this WMT shared task.
Augmenting the training set with artificially synthesized data is one of the popular and effective approaches for coping with this challenge.
It was first used to improve the quality of \acrshort{nmt} systems \cite{sennrich-etal-2016-improving} and then it was applied to the \acrshort{ape} task \cite{junczys-dowmunt-grundkiewicz-2016-log}.
This approach, however, showed limited success on automatically post editing the high quality translations of \acrshort{ape} systems.

Transfer learning is another solution to deal with data sparsity in such tasks.
It is based on the assumption that the knowledge extracted from other well-resourced tasks can be transferred to the new tasks/domains.
Recently, large models pre-trained on multiple tasks with vast amounts of data, for instance BERT and MT-DNN~\cite{devlin2018bert, liu2019mt-dnn}, have obtained state-of-the-art results when fine-tuned over a small set of training samples. 
Following \newcite{apebert19}, in this paper we use BERT~\cite{devlin2018bert} within the encoder-decoder framework (\S\ref{approach:bed}) and formulate the task of \acrlong{ape} as generating \texttt{pe} which is (possibly) the modified version of \texttt{mt} given the original source sentence \texttt{src}. 
As discussed in \S\ref{approach:bed}, instead of using multi-encoder architecture, in this work we concatenate the \texttt{src} and \texttt{mt} with the BERT special token (i.e. \texttt{[SEP]} and feed them to our single encoder.

We also introduce the \textit{conservativeness penalty}, a simple yet effective mechanism that controls the freedom of our \ac{ape} in modifying the given MT output. As we show in \S\ref{approach:conserv}, in the cases where the automatic translations are of high quality, this factor forces the \ac{ape} system to do less modifications, hence avoids the well-known problem of over-correction. 

Finally, we augmented our original in-domain training data with a synthetic corpus which contains around $3M$ \texttt{<src,mt,pe>} triplets (\S\ref{exp:data}). 
As discussed in \S\ref{results}, our system is able to improve significantly the MT outputs by $-0.78$ TER~\cite{snover2006study} and $+1.23$ BLEU~\cite{papineni2002bleu}, achieving an ex-aequo first-place in the English-German track.

\section{Approach}
In this section we describe the main features of our \ac{ape} system: the \textit{BERT-based encoder-decoder} (BED) and the \textit{conservativeness penalty}.

\subsection{BERT-based encoder-decoder}
\label{approach:bed}
\begin{figure}[tb!]
    \centering
    \includegraphics[width=\linewidth]{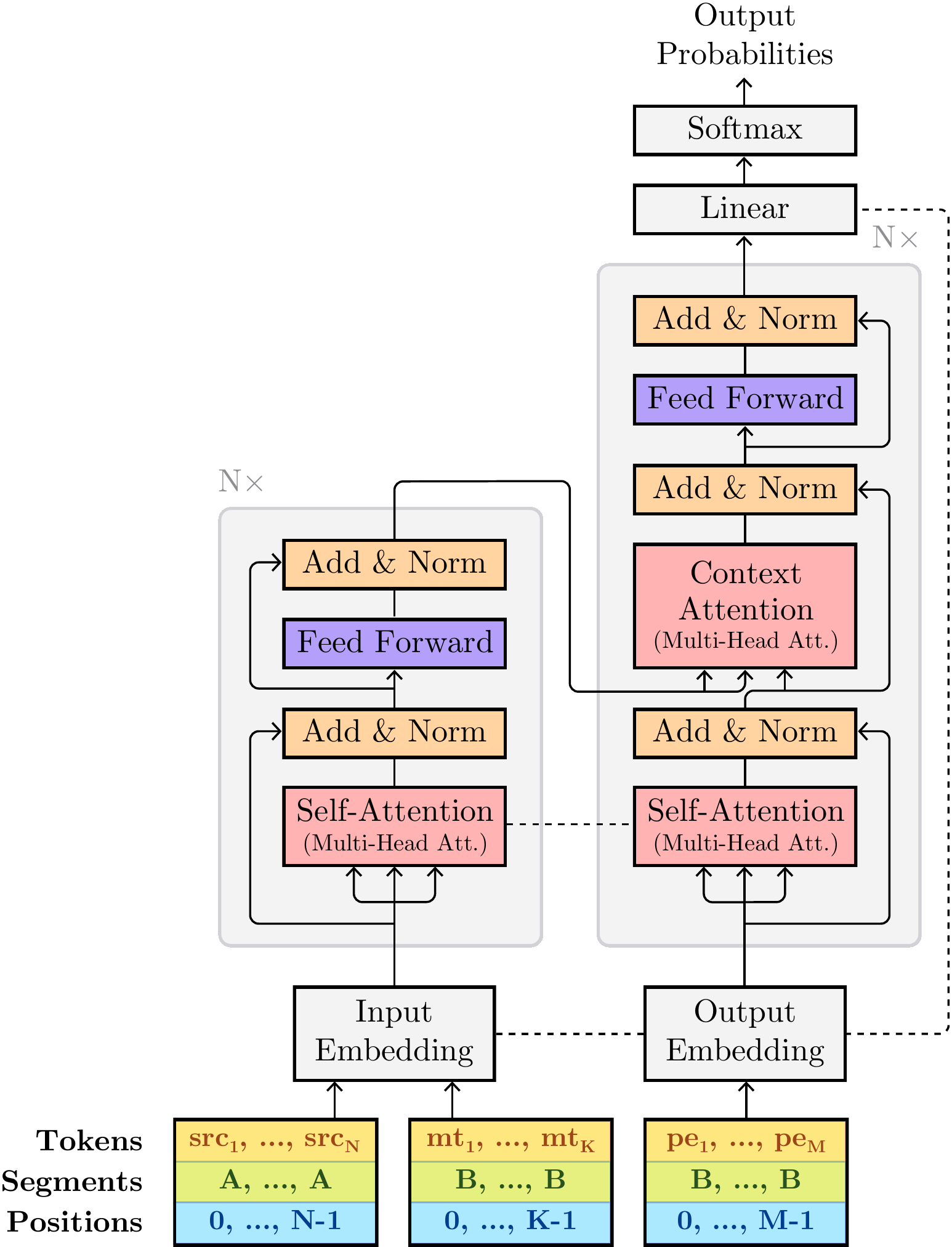}
    \caption{BERT encoder decoder, taken from~\newcite{apebert19}.}
    \label{fig:bert-encoder-decoder}
\end{figure}

Following \cite{apebert19} we adapt the BERT model to the \acrshort{ape} task by integrating the model in an encoder-decoder architecture.
To this aim we use a single BERT encoder to obtain a joint representation of the ~\texttt{src} and ~\texttt{mt} sentence and a BERT-based decoder where the multi-head context attention block is initialized with the weights of the self-attention block. Both the encoder and the decoder are initialized with the pre-trained weights of the multilingual BERT\footnote{\url{https://github.com/google-research/bert}}~\cite{bert:2018}.
Figure~\ref{fig:bert-encoder-decoder} depicts our BED model. 

Instead of using multiple encoders to separately encode \texttt{src} and \texttt{mt}, we use BERT pre-training scheme, where the two strings after being concatenated by the  \texttt{[SEP]} special symbol are fed to the single encoder.
We treat these sentences as \texttt{sentenceA} and \texttt{sentenceB} in \cite{bert:2018} and assign different segment embeddings to each of them.
This emulates a similar setting to ~\cite{DBLP:conf/wmt/Junczys-Dowmunt18a} where a dual-source encoder with shared parameters is used to encode both input strings.

On the target side, following \cite{apebert19} we use a single decoder where the context attention block is initialized with the self attention weights, and all the weights of the self-attention are shared with the respective self-attention weights in the encoder.

\subsection{Conservativeness penalty}
\label{approach:conserv}
With domain specific~\ac{nmt} systems making relatively few translation errors, \ac{ape} systems face new challenges.
This means more careful decisions have to be made by the \ac{ape} system, making the least possible edits to the raw \texttt{mt}. 
To this aim, we introduce our ``conservativeness'' penalty developed on the \textit{post editing penalty} proposed by \cite{junczys-dowmunt-grundkiewicz-2016-log}.
It is a simple yet effective method to penalize/reward hypotheses in the beam, at inference time, that diverge far from the original input. 

More formally, let $V$ be the source and target 
%shared 
vocabulary. 
We define $V_{c}=\{V_{src}\cup~V_{mt}\}$ as the conservative tokens of an APE triplet, where $V_{src},V_{mt}\subset V$ are the \texttt{src} and \texttt{mt} tokens, respectively. 
For the sake of argument we define $V_c$ for decoding a single \ac{ape} triplet, which can be generalized to batch decoding with $V_c$ defined for each batch element.
Given the $|V|$ sized vector of candidates $h_{t}$ at each decoding step $t$, we modify the score/probability of each candidate $v$ as: 
\begin{equation}
 h_{t}(v) = 
  \begin{cases}
    h_{t}(v) - c       & \quad \text{if } v \in V \setminus V_{c}\\
    h_{t}(v)  & \quad \text{otherwise}
  \end{cases}
  \label{eq:conservativeness}
\end{equation}
%where $c \in (-\infty,+\infty)$ is the conservativeness penalty.
where $c$ is the conservativeness penalty, penalizing (or rewarding for negative values) all tokens of $V$ not present in $V_c$.
Note that, this penalty can be applied to either the raw non-normalized outputs of the model (logit) or the final probabilities (log probabilities).

As the log probabilities and logit scores have different bounds of $(-\infty,0)$ and $(-\infty,+\infty)$, respectively, $c$ is set accordingly.
Hence, for positive values of conservativeness the aim is to avoid picking tokens not in the \texttt{src} and \texttt{mt}, thus, limiting the number of corrections. On the other hand, negative values enable over correction. 

Moreover, in order to apply the penalty in the log probabilities, there are some considerations to take into account as we don't renormalize after the transformation. For positive values, the factor lowers the probability of all non conservative tokens, either increasing the confidence of an already picked conservative token, or favouring these tokens that are close to the best candidate -- thus being closer to scores rather than probabilities.
In contrast, negative penalties might require carefully selected values and truncating at the upper boundary -- we did not experiment with negative values in this work, however the Quality Estimation shared task winning system used an APE-QE system with negative conservativeness~\cite{unbabelqewmt}.

In contrast with \citeauthor{junczys-dowmunt-grundkiewicz-2016-log}, our model takes into account both \texttt{src} and \texttt{mt}, allowing to copy either of them directly.
This is beneficial to handle proper nouns as they should be preserved in the post edition without any modification.
Moreover, instead of setting the penalty as a fixed value of $-1$, we define it as a hyperparameter which enables a more dynamic control of our model's post-editions to the \texttt{mt} input.

\begin{table*}[ht]
\centering

\begin{tabular}{|l|c|c|c|c|}
\hline
\multicolumn{1}{|c|}{System} & Beam & $\downarrow$w/o c       & $\downarrow$best c                                        & $\downarrow$worst c         \\ \hline
MT Baseline                  & -    & 15.08      & -                                             & -               \\ \hline
\multirow{2}{*}{BED}         & 4    & 15.65      & -                                             & -               \\ \cline{2-5} 
                             & 6    & 15.61      & -                                             & -               \\ \hline
\multirow{2}{*}{+ logprobs}  & 4    & \textbf{-} & \textbf{14.84 ($c=1.5$)}                      & 15.06 ($c=2.3$) \\ \cline{2-5} 
                             & 6    & \textbf{-} & \multicolumn{1}{l|}{\textbf{14.87 ($c=1.5$)}} & 15.01 ($c=2.5$)  \\ \hline
\multirow{2}{*}{+ logits}    & 4    & -          & \multicolumn{1}{l|}{15.03 ($c=1.7$)}          & 15.25 ($c=0.9$) \\ \cline{2-5} 
                             & 6    & -          & \multicolumn{1}{l|}{15.05 ($c=1.7$)}          & 15.23 ($c=0.9$) \\ \hline
\end{tabular}

\caption{TER scores of the baseline NMT system and our BERT encoder-decoder \texttt{ape} model.
The columns ``w/o c'', ``best c'', and ``worst c'' presents the scores of our system without the conservativeness penalty, with the best and the worst conservativeness penalty settings on our dev corpus, respectively.
``logprobs'' and ``logits'' refer, respectively, to the state where we apply the conservativeness factor (see ~\S\ref{approach:conserv})}
\label{tab:grid-cons}
\end{table*}

\section{Experiments}
\subsection{Data}
\label{exp:data}
This year for the English-German language pair the participants were provided an in-domain training set and the eSCAPE corpus, an artificially synthesized generic training corpus for \ac{ape} \cite{negri2018escape}.
In addition to these corpora, they were allowed to use any additional data to train their systems.
Considering this, and the fact that the in-domain training set belongs to the IT domain, we decided to use our own synthetic training corpus.
Thus, we trained our models on a combination of the in-domain data released by the \ac{ape} task organizers and this synthetic dataset.

\textbf{In-domain training set:} we use the $13k$ triplets of \texttt{<src,mt,pe>} in the IT domain without any preprocessing as they are already preprocessed by the shared task organizers. 
Despite the previous year where the \texttt{mt} side was generated either by a phrase-based or a neural \ac{mt} system, this year all the source sentences were translated only by a neural \ac{mt} system unknown to the participants.

\textbf{Synthetic training set:} instead of the eSCAPE corpus provided by the organizers we created our own synthetic corpus using the parallel data provided by the Quality Estimation shared task\footnote{Dataset can be found under Additional Resouces at \url{http://www.statmt.org/wmt19/qe-task.html}}.
We found this corpus closer to the IT domain which is the target domain of the \ac{ape} task.
To create this corpus we performed the following steps:
\begin{enumerate}
    \item Split the corpus into $5$ folds $f_i$.
    \item Use OpenNMT~\cite{opennmt} to train 5 LSTM based translation models, one model $\mathcal{M}_i$ for every subset created by removing fold $f_i$ from the training data.
    \item Translate each fold $f_i$ using the translation Model $\mathcal{M}_i$.
    \item Join the translations to get an unbiased machine translated version of the full corpus.
    \item Remove empty lines.
\end{enumerate}

The final corpus has $3.3M$ triplets.
We then oversampled the in-domain training data 20 times ~\cite{DBLP:conf/wmt/Junczys-Dowmunt18a} and used them together with our synthetic data to train our models.

\subsection{BED training}
We follow \citeauthor{apebert19} for training our BERT-based Encoder-Decoder \ac{ape} models.
In particular, we set the learning rate to $5e^{-5}$ and use $bertadam$ optimizer to perform $200k$ steps from which $20k$ are warmup steps.
We set the effective batch size to $2048$ tokens.
Furthermore, we also use a shared matrix for the input and output token embedddings and the projection layer \cite{press2017using}.
Finally, we share the self-attention weights between the encoder and the decoder and initialize the multi-head attention of the decoder with the self-attention weights of the encoder.

Similarly to \newcite{junczys2018microsoft}, we apply a data weighting strategy during training. However, we use a different weighting approach, where each sample $s_i$ is assigned a weight, $w_{s_i}$, defined as $1-TER(s_i)$.
This results in assigning higher weights to the samples with less \ac{mt} errors and vice versa, which might sound counter intuitive since in the APE task the goal is to learn more from the samples with larger number of errors.
However, in this task, where the translations are provided by strong \ac{nmt} systems with very small number of errors, our \ac{ape} system needs to be conservative and learn to perform limited number of modifications to the \texttt{mt}.

\subsection{BED decoding}
In the decoding step we perform the standard beam decoding with our conservativeness factor.
We fine tuned the this factor on the dev set provided by the organizers.
Furthermore, in our experiments we set restrict the search to $c \in [0, +5]$ and use beam sizes of 4 and 6.
In our preliminary experiments larger beam sizes didn't help to improve the performance further.
Finally, we used the evaluation script available on the website to access the performance of our model.

\section{Results and discussion}
\label{results}
In our preliminary experiments we noticed that using the pure BED model does not improve the quality of the translations provided by strong \ac{nmt} systems. 
As Table \ref{tab:grid-cons} shows, it actually degrades the performance by $-0.57$ TER scores.
Although the scores in \citeauthor{apebert19} are actually closer to the baseline, we find that using the BED model only, without controlling the conservativeness to the original \ac{mt} can lead to baseline level scores (on dev).
Hence, we applied different conservativeness penalties during the beam decoding and as the results in Table\ref{tab:grid-cons} show, different values for this hyperparameter significantly changes the performance of our model. 
% However, in contrast with~\citeauthor{junczys-dowmunt-grundkiewicz-2016-log}'s gains for the SMT task, we see more marginal improvements with our penalty. Table~\ref{tab:grid-cons} depicts the effect of the conservativeness penalty.
For the sake of compactness, here we present only the best (i.e. \texttt{best c}) and worst (i.e. \texttt{worst c}) scores by our model, to compare the effect of this factor. 
 
Furthermore, intuitively, logits stands as the best candidate to apply the penalty, not only it was done in a similar fashion previously~\cite{DBLP:conf/wmt/Junczys-Dowmunt18a}, but also, after the normalization of the weights, the conservative tokens should have large peaks while having a stable behaviour. However, we achieved our best scores with penalties over the log probabilities, suggesting pruning hypothesis directly after normalizing the logits leads to more conservative outputs. 
Nonetheless, we leave as future work further investigations on the impact of pruning before and after normalizing the logits, as well as exploring renormalization of the log probabilities.
Finally, we hypothesize that not only our BED model but also other \ac{ape} models could benefit from the conservativeness penalty.
We, however, leave it to be explored in future work.

Regarding the performance of our model on the official test set, as the results of Table \ref{res:oursys} show, we outperform last year's winning systems by almost $-0.4$ TER and $+0.5$ BLEU, which for strong performing~\acrshort{nmt} systems is significant. In addition, our submission ranks first in the official results~\footnote{Available at \url{http://www.statmt.org/wmt19/ape-task.html} under \textit{Results}.}, \textit{ex aequo} with 3 other systems. Table~\ref{tab:fullres} depicts the official results of the shared task, considering only the best submission of each team.

\begin{table}[htbp]
\centering
%\small
\begin{tabular}{l|c|c|}
\cline{2-3}
                                    & $\downarrow$TER & $\uparrow$BLEU                       \\ \hline
\multicolumn{1}{|l|}{Baseline}      & 16.84 & 74.73                      \\ \hline\hline
%\multicolumn{1}{|l|}{\citeauthor{DBLP:conf/wmt/Junczys-Dowmunt18a}}      & 16.84 & 74.73                      \\ \hline
\multicolumn{1}{|l|}{\cite{tebbifakhr-etal-2018-multi}}      & 16.46 & 75.53  \\ \hline\hline
\multicolumn{1}{|l|}{Primary}       & \bf{16.08} & \bf{75.96}            \\ \hline
\multicolumn{1}{|l|}{Contrastive} & 16.21 & 75.70                      \\ \hline
\end{tabular}
\caption{Submission at the WMT APE shared task.}
\label{res:oursys}
\end{table}

Although in this paper we did not present an ablation analysis (due to time constraints), we hypothesize that three BED training and decoding techniques used in this work were influential on the final result obtained for this task:
i) the synthetic training corpus contains more IT domain samples than the generic eSCAPE corpus, making it a suitable dataset to train \ac{ape} systems for this domain;
ii) the data weighting mechanism enforces the system to be more conservative and learn fewer edits which is crucial for strong specialized \ac{nmt} engines,
and, finally, iii) the conservativeness factor is crucial to avoid the well-known problem of over-correction posed generally by \ac{ape} systems over the high quality \ac{nmt} outputs, guaranteeing faithfulness to the original~\acrshort{mt}.

\begin{table}
\centering
\begin{tabular}{lcc}
\hline
System     & $\downarrow$Ter   & $\uparrow$BLEU  \\ \hline
\textbf{Ours (Unbabel)}       & \textbf{16.06}$^\star$ & 75.96 \\ %\hline
POSTECH    & $16.11^\star$ & \textbf{76.22} \\% \hline
USSAR DFKI & $16.15^\star$ & 75.75 \\% \hline
FBK        & $16.37^\star$ & 75.71 \\% \hline
UdS MTL    & 16.77 & 75.03 \\% \hline
IC USFD    & 16.78 & 74.88 \\% \hline
Baseline   & 16.84 & 74.73 \\% \hline
ADAP DCU   & 17.07 & 74.30 \\ \hline
\end{tabular}
\caption{APE Results as provided by the shared task organizers. We only present the best score of each team. $\star$ indicates not statistically significantly different, \textit{ex aequo}.}
\label{tab:fullres}
\end{table}

\section{Conclusion}\label{conclusion}
We presented Unbabel's submissions to the APE shared task at WMT 2019 for the English-German language pair. 
Our model uses the BERT pre-trained language model within the encoder-decoder framework and applies a conservative factor to control the faithfulness of \ac{ape} system to the original input stream.

The result of the official evaluation show that our system is able to effectively detect and correct the few errors made by the strong \ac{nmt} system, improving the score by $-0.8$ and $+1.2$ in terms of TER and BLEU, respectively. 

Finally, using~\acrshort{ape} to improve strong in-domain~\acrlong{nmt} systems is increasingly more challenging, and ideally the editing system will tend to perform less and less modifications of the raw~\texttt{mt}. In line with~\citeauthor{DBLP:conf/wmt/Junczys-Dowmunt18a}'s suggestion, studying how to apply~\acrshort{ape} to engines in generic data (domain agnostic) can be a more challenging task, as it would require more robustness and generalization of the~\acrshort{ape} system.

\section*{Acknowledgments}
The authors would like to thank the anonymous reviewers for the feedback. Moreover, we would like to thank Ant\'{o}nio G\'{o}is, F\'{a}bio Kepler, and Miguel Vera for the fruitful discussions and help. We would also like to thank the support provided by the EU in the context of the PT2020 project (contracts 027767 and 038510), by the European Research Council (ERC StG DeepSPIN 758969), and by the Funda\c{c}\~{a}o para a Ci\^{e}ncia e Tecnologia through contract UID/EEA/50008/2019.

\bibliography{acl2019}

\begin{thebibliography}{20}
\expandafter\ifx\csname natexlab\endcsname\relax\def\natexlab#1{#1}\fi

\bibitem[{Bojar et~al.(2016)Bojar, Chatterjee, Federmann, Graham, Haddow, Huck,
  Jimeno~Yepes, Koehn, Logacheva, Monz, Negri, Neveol, Neves, Popel, Post,
  Rubino, Scarton, Specia, Turchi, Verspoor, and Zampieri}]{wmt-findings:2016}
Ond{\v{r}}ej Bojar, Rajen Chatterjee, Christian Federmann, Yvette Graham, Barry
  Haddow, Matthias Huck, Antonio Jimeno~Yepes, Philipp Koehn, Varvara
  Logacheva, Christof Monz, Matteo Negri, Aurelie Neveol, Mariana Neves, Martin
  Popel, Matt Post, Raphael Rubino, Carolina Scarton, Lucia Specia, Marco
  Turchi, Karin Verspoor, and Marcos Zampieri. 2016.
\newblock \href {https://doi.org/10.18653/v1/W16-2301} {Findings of the 2016
  conference on machine translation}.
\newblock In \emph{Proceedings of the First Conference on Machine Translation},
  pages 131--198, Berlin, Germany. Association for Computational Linguistics.

\bibitem[{Bojar et~al.(2015)Bojar, Chatterjee, Federmann, Haddow, Huck, Hokamp,
  Koehn, Logacheva, Monz, Negri, Post, Scarton, Specia, and
  Turchi}]{wmt-findings:2015}
Ond{\v{r}}ej Bojar, Rajen Chatterjee, Christian Federmann, Barry Haddow,
  Matthias Huck, Chris Hokamp, Philipp Koehn, Varvara Logacheva, Christof Monz,
  Matteo Negri, Matt Post, Carolina Scarton, Lucia Specia, and Marco Turchi.
  2015.
\newblock \href {https://doi.org/10.18653/v1/W15-3001} {Findings of the 2015
  workshop on statistical machine translation}.
\newblock In \emph{Proceedings of the Tenth Workshop on Statistical Machine
  Translation}, pages 1--46, Lisbon, Portugal. Association for Computational
  Linguistics.

\bibitem[{Chatterjee et~al.(2018{\natexlab{a}})Chatterjee, Negri, Rubino, and
  Turchi}]{wmt-findings:2017}
Rajen Chatterjee, Matteo Negri, Raphael Rubino, and Marco Turchi.
  2018{\natexlab{a}}.
\newblock \href {https://www.aclweb.org/anthology/W18-6452} {Findings of the
  {WMT} 2018 shared task on automatic post-editing}.
\newblock In \emph{Proceedings of the Third Conference on Machine Translation:
  Shared Task Papers}, pages 710--725, Belgium, Brussels. Association for
  Computational Linguistics.

\bibitem[{Chatterjee et~al.(2018{\natexlab{b}})Chatterjee, Negri, Rubino, and
  Turchi}]{wmt-findings:2018}
Rajen Chatterjee, Matteo Negri, Raphael Rubino, and Marco Turchi.
  2018{\natexlab{b}}.
\newblock \href {https://www.aclweb.org/anthology/W18-6452} {Findings of the
  {WMT} 2018 shared task on automatic post-editing}.
\newblock In \emph{Proceedings of the Third Conference on Machine Translation:
  Shared Task Papers}, pages 710--725, Belgium, Brussels. Association for
  Computational Linguistics.

\bibitem[{Correia and Martins(2019)}]{apebert19}
Gon\c{c}alo Correia and Andr\'{e} Martins. 2019.
\newblock A simple and effective approach to automatic post-editing with
  transfer learning.
\newblock In \emph{Proceedings of the 57th annual meeting on association for
  computational linguistics}. Association for Computational Linguistics.

\bibitem[{Devlin et~al.(2018{\natexlab{a}})Devlin, Chang, Lee, and
  Toutanova}]{devlin2018bert}
Jacob Devlin, Ming-Wei Chang, Kenton Lee, and Kristina Toutanova.
  2018{\natexlab{a}}.
\newblock Bert: Pre-training of deep bidirectional transformers for language
  understanding.
\newblock \emph{arXiv preprint arXiv:1810.04805}.

\bibitem[{Devlin et~al.(2018{\natexlab{b}})Devlin, Chang, Lee, and
  Toutanova}]{bert:2018}
Jacob Devlin, Ming-Wei Chang, Kenton Lee, and Kristina Toutanova.
  2018{\natexlab{b}}.
\newblock Bert: Pre-training of deep bidirectional transformers for language
  understanding.
\newblock \emph{arXiv preprint arXiv:1810.04805}.

\bibitem[{Junczys-Dowmunt(2018)}]{junczys2018microsoft}
Marcin Junczys-Dowmunt. 2018.
\newblock Microsoft's submission to the wmt2018 news translation task: How i
  learned to stop worrying and love the data.
\newblock In \emph{Proceedings of the Third Conference on Machine Translation:
  Shared Task Papers}, pages 425--430.

\bibitem[{Junczys-Dowmunt and
  Grundkiewicz(2016)}]{junczys-dowmunt-grundkiewicz-2016-log}
Marcin Junczys-Dowmunt and Roman Grundkiewicz. 2016.
\newblock \href {https://doi.org/10.18653/v1/W16-2378} {Log-linear combinations
  of monolingual and bilingual neural machine translation models for automatic
  post-editing}.
\newblock In \emph{Proceedings of the First Conference on Machine Translation},
  pages 751--758, Berlin, Germany. Association for Computational Linguistics.

\bibitem[{Junczys{-}Dowmunt and
  Grundkiewicz(2018)}]{DBLP:conf/wmt/Junczys-Dowmunt18a}
Marcin Junczys{-}Dowmunt and Roman Grundkiewicz. 2018.
\newblock \href {https://aclanthology.info/papers/W18-6467/w18-6467}
  {{MS}-{UEdin} submission to the {WMT2018} {APE} shared task: Dual-source
  transformer for automatic post-editing}.
\newblock In \emph{Proceedings of the Third Conference on Machine Translation:
  Shared Task Papers, {WMT} 2018, Belgium, Brussels, October 31 - November 1,
  2018}, pages 822--826.

\bibitem[{Kepler et~al.(2019)Kepler, Tr\'{e}nous, Treviso, Vera, G\'{o}is,
  Farajian, V.~Lopes, and F.~T.~Martins}]{unbabelqewmt}
Fabio Kepler, Jonay Tr\'{e}nous, Marcos Treviso, Miguel Vera, Ant\'{o}nio
  G\'{o}is, M.~Amin Farajian, Ant\'{o}nio V.~Lopes, and Andr\'{e}
  F.~T.~Martins. 2019.
\newblock Unbabel’s participation in the wmt19 translation quality estimation
  shared task.
\newblock In \emph{Proceedings of the Fourth Conference on Machine
  Translation}, Florence, Italy. Association for Computational Linguistics.

\bibitem[{Klein et~al.(2017)Klein, Kim, Deng, Senellart, and Rush}]{opennmt}
Guillaume Klein, Yoon Kim, Yuntian Deng, Jean Senellart, and Alexander~M. Rush.
  2017.
\newblock \href {https://doi.org/10.18653/v1/P17-4012} {Open{NMT}: Open-source
  toolkit for neural machine translation}.
\newblock In \emph{Proc. ACL}.

\bibitem[{Knight and Chander(1994)}]{Knight:1994:APD:2891730.2891850}
Kevin Knight and Ishwar Chander. 1994.
\newblock \href {http://dl.acm.org/citation.cfm?id=2891730.2891850} {Automated
  postediting of documents}.
\newblock In \emph{Proceedings of the Twelfth AAAI National Conference on
  Artificial Intelligence}, AAAI'94, pages 779--784. AAAI Press.

\bibitem[{Liu et~al.(2019)Liu, He, Chen, and Gao}]{liu2019mt-dnn}
Xiaodong Liu, Pengcheng He, Weizhu Chen, and Jianfeng Gao. 2019.
\newblock Multi-task deep neural networks for natural language understanding.
\newblock \emph{arXiv preprint arXiv:1901.11504}.

\bibitem[{Negri et~al.(2018)Negri, Turchi, Chatterjee, and
  Bertoldi}]{negri2018escape}
Matteo Negri, Marco Turchi, Rajen Chatterjee, and Nicola Bertoldi. 2018.
\newblock escape: a large-scale synthetic corpus for automatic post-editing.
\newblock In \emph{LREC 2018, Eleventh International Conference on Language
  Resources and Evaluation}, pages 24--30. European Language Resources
  Association (ELRA).

\bibitem[{Papineni et~al.(2002)Papineni, Roukos, Ward, and
  Zhu}]{papineni2002bleu}
Kishore Papineni, Salim Roukos, Todd Ward, and Wei-Jing Zhu. 2002.
\newblock Bleu: a method for automatic evaluation of machine translation.
\newblock In \emph{Proceedings of the 40th annual meeting on association for
  computational linguistics}, pages 311--318. Association for Computational
  Linguistics.

\bibitem[{Press and Wolf(2017)}]{press2017using}
Ofir Press and Lior Wolf. 2017.
\newblock Using the output embedding to improve language models.
\newblock In \emph{Proceedings of the 15th Conference of the European Chapter
  of the Association for Computational Linguistics: Volume 2, Short Papers},
  pages 157--163, Valencia, Spain. Association for Computational Linguistics.

\bibitem[{Sennrich et~al.(2016)Sennrich, Haddow, and
  Birch}]{sennrich-etal-2016-improving}
Rico Sennrich, Barry Haddow, and Alexandra Birch. 2016.
\newblock \href {https://doi.org/10.18653/v1/P16-1009} {Improving neural
  machine translation models with monolingual data}.
\newblock In \emph{Proceedings of the 54th Annual Meeting of the Association
  for Computational Linguistics (Volume 1: Long Papers)}, pages 86--96, Berlin,
  Germany. Association for Computational Linguistics.

\bibitem[{Snover et~al.(2016)Snover, Dorr, Schwartz, Micciulla, and
  Makhoul}]{snover2006study}
Matthew Snover, Bonnie Dorr, Richard Schwartz, Linnea Micciulla, and John
  Makhoul. 2016.
\newblock A study of translation edit rate with targeted human annotation.
\newblock In \emph{Proceedings of association for machine translation in the
  Americas}, pages Vol. 200, No. 6.

\bibitem[{Tebbifakhr et~al.(2018)Tebbifakhr, Agrawal, Negri, and
  Turchi}]{tebbifakhr-etal-2018-multi}
Amirhossein Tebbifakhr, Ruchit Agrawal, Matteo Negri, and Marco Turchi. 2018.
\newblock \href {https://www.aclweb.org/anthology/W18-6471} {Multi-source
  transformer with combined losses for automatic post editing}.
\newblock In \emph{Proceedings of the Third Conference on Machine Translation:
  Shared Task Papers}, pages 846--852, Belgium, Brussels. Association for
  Computational Linguistics.

\end{thebibliography}
\bibliographystyle{acl_natbib}

\end{document}